\documentclass[sigconf]{acmart}
\renewcommand\footnotetextcopyrightpermission[1]{} % removes footnote with conference information in first column
\usepackage{booktabs} % For formal tables
\usepackage{amsmath,amsfonts,amssymb,epsfig,epstopdf,url,array}

\usepackage{amsthm}
\usepackage{color,soul}
\usepackage{algorithmic}
\usepackage{balance}
\usepackage[ruled]{algorithm2e}
\usepackage{stmaryrd}
\usepackage{centernot}
\usepackage{tikz}
\usepackage{booktabs}
\usepackage{pdfpages}
\usepackage{graphicx}
\usepackage{subfigure}
\usepackage[font=small,skip=8pt]{caption}
\usepackage{tabularx}
\usepackage{comment}
\usepackage{multirow}
\usepackage{multicol}
\usepackage{bigstrut}
\usepackage{pdflscape}
\usepackage{rotating}
\usepackage{float}
\usepackage{lipsum}

\usepackage{enumitem}
\usetikzlibrary{plotmarks,shapes,arrows,chains,hobby,backgrounds,calc,trees}
\usepackage{makecell}
\usepackage{dblfloatfix}    % To enable figures at the bottom of page
\usepackage{pdfcomment} %\pdfmarkupcomment[markup=Highlight,color=yellow]{\emph{and not accurate and efficient}}{comment}
\usepackage{etoolbox}
\usepackage{colortbl}

\definecolor{lightgray}{gray}{0.9}
\theoremstyle{definition}

\theoremstyle{definition}

\theoremstyle{definition}
\newtheorem{defn}{Definition}

\setcopyright{none}
\makeatletter
\def\@copyrightspace{\relax}
\makeatother

\begin{document}
\title{An Integer Programming Model for Binary Knapsack Problem with Value-Related Dependencies among Elements}
%\title{An Integer Linear Programming Model \\for Maximizing the Lower Bound of Utility}

\author{Davoud Mougouei}
%\authornote{Corresponding Author}
\affiliation{%
  \institution{School of Computer Science, Engineering, and Mathematics
  	Flinders University}
  \city{Adelaide, Australia} 
}
\email{davoud.mougouei@flinders.edu.au}

\author{David M. W. Powers}
\affiliation{%
	\institution{School of Computer Science, Engineering, and Mathematics
		Flinders University}
	\city{Adelaide, Australia} 
}
\email{david.powers@flinders.edu.au}

\author{Asghar Moeini}
\affiliation{%
	\institution{School of Computer Science, Engineering, and Mathematics
		Flinders University}
	\city{Adelaide, Australia} 
}
\email{asghar.moeini@flinders.edu.au}

% The default list of authors is too long for headers}
\renewcommand{\shortauthors}{D. Mougouei et al.}

\begin{abstract}
\textit{Binary Knapsack Problem} (BKP) is to select a subset of an element (item) set with the highest value while keeping the total weight within the capacity of the knapsack. This paper presents an integer programming model for a variation of BKP where the value of each element may depend on selecting or ignoring other elements. Strengths of such \textit{Value-Related Dependencies} are assumed to be imprecise and hard to specify. To capture this imprecision, we have proposed modeling value-related dependencies using fuzzy graphs and their algebraic structure.
\end{abstract}

%
% The code below should be generated by the tool at
% http://dl.acm.org/ccs.cfm
% Please copy and paste the code instead of the example below. 
%
%\begin{CCSXML}
%<ccs2012>
% <concept>
%  <concept_id>10010520.10010553.10010562</concept_id>
%  <concept_desc>Computer systems organization~Embedded systems</concept_desc>
%  <concept_significance>500</concept_significance>
% </concept>
% <concept>
%  <concept_id>10010520.10010575.10010755</concept_id>
%  <concept_desc>Computer systems organization~Redundancy</concept_desc>
%  <concept_significance>300</concept_significance>
% </concept>
% <concept>
%  <concept_id>10010520.10010553.10010554</concept_id>
%  <concept_desc>Computer systems organization~Robotics</concept_desc>
%  <concept_significance>100</concept_significance>
% </concept>
% <concept>
%  <concept_id>10003033.10003083.10003095</concept_id>
%  <concept_desc>Networks~Network reliability</concept_desc>
%  <concept_significance>100</concept_significance>
% </concept>
%</ccs2012>  
%\end{CCSXML}
%
%\ccsdesc[500]{Computer systems organization~Embedded systems}
%\ccsdesc[300]{Computer systems organization~Redundancy}
%\ccsdesc{Computer systems organization~Robotics}
%\ccsdesc[100]{Networks~Network reliability}

% We no longer use \terms command
%\terms{Theory}

\keywords{Binary Knapsack Problem, Integer Programming, Dependency, Value}

\maketitle

\section{Modeling Dependencies}
\label{sec_modleing}

Fuzzy graphs have demonstrated to properly capture imprecision of real world problems~\cite{mordeson_applications_2000,mougouei2017dasrp}. Hence, we have used algebraic structure of fuzzy graphs for capturing the imprecision associated with strengths of value-related dependencies. We have specially modified the classical definition of fuzzy graphs in order to consider not only the strengths but also the qualities (positive or negative)~\cite{mougouei2016factoring,mougouei2017dasrp,mougouei2017preference} of value-related dependencies (Definition~\ref{def_RDG}).

\begin{defn}
\label{def_RDG}
\textit{Value Dependency Graph} (VDG). A VDG is a signed directed fuzzy graph~\cite{Wasserman1994} $G=(E,\sigma,\rho)$ in which a non-empty set of elements $E:\{e_1,...,e_n\}$ constitute the graph nodes. Also, the qualitative function $\sigma: E\times E\rightarrow \{+,-,\pm\}$ and the membership function $\rho: E\times E\rightarrow [0,1]$ denote qualities and strengths of explicit value-related dependencies receptively. As such, a pair of elements $(e_i,e_j)$ with $\rho_{i,j}\neq 0$ and $\sigma_{i,j}\neq \pm$ denotes an explicit value-related dependency from $e_i$ to $e_j$. It is clear that we have $\rho_{i,j}=0$ if the value of an element $e_i$ is not explicitly influenced by selecting or ignoring $e_j$. In such cases we have $\sigma_{i,j}=\pm$ where $\pm$ denotes the quality of $(e_i,e_j)$ is \text{non-specified}. 
\end{defn}

\begin{defn}
	\label{def_RDG_valuedepndencies}
	\textit{Value-Related Dependencies}. 
	A value-related dependency in a VDG $G=(E,\sigma,\rho)$ is defined as a sequence of elements $d_i:\big(e(1),...,e(k)\big)$ such that for each $e(j)$ in $d_i$, $2 \leq j \leq k$, we have $\rho_{j-1,j} \neq 0$. A consecutive pair $\big(e(j-1),e(j)\big)$ specifies an explicit value-related dependency from $e(j-1)$ to $e(j)$. 
\end{defn}

%Based on the Definition \ref{def_RDG}, an explicit value-related dependency from an element $e_i$ to $e_j$ is specified by $(e_i,e_j)$ where $\rho_{i,j}\neq 0$ and $\sigma_{i,j}$ specifies the quality (positivity/negativity) of $(e_i,e_j)$. Definition~\ref{def_RDG_valuedepndencies} however, includes both explicit and implicit value-related dependencies.   

The strength of a value-related dependency $d_i:\big(e(1),...,e(k)\big)$ is calculated by (\ref{Eq_RDG_strength}). That is, the strength of a value-related dependency equals to the strength of the weakest of all the $k-1$ explicit value-related dependencies along the path. $\wedge$ denote Zadeh's~\cite{zadeh_fuzzysets_1965} fuzzy AND operator that is taking minimum over operands. 

\begin{eqnarray}
\label{Eq_RDG_strength}
\forall d_i:\big(e(1),...,e(k)\big), \quad \rho(d_i) = \bigwedge_{j=2}^{k}\rho_{j-1,j}
\end{eqnarray}

The quality of a value-related dependency $d_i:\big(e(1),...,e(k)\big)$ on the other hand, is calculated through employing qualitative sequential inference~\cite{de1984qualitative,wellman1990formulation,kusiak_1995_dependency} as given by (\ref{Eq_RDG_quality}). %Table~\ref{table_inference} shows how qualitative serial inference gives the quality of value-related dependencies.  

\begin{eqnarray}
\label{Eq_RDG_quality}
\forall d_i:\big(e(1),...,e(k)\big),\quad \sigma(d_i) = \prod_{j=2}^{k}\sigma_{j-1,j}
\end{eqnarray}
\vspace{0.1em}
%---------------------------------------------------------------
%\begin{table}[!htb]
%	\caption{Qualitative Serial Inference in VDGs.}
%	\label{table_inference}
%	\centering
%	\input{tables/table_inference}
%\end{table}

Let $D=\{d_1,d_2,..., d_m\}$ be the set of all explicit and implicit value-related dependencies from $e_i \in E$ to $e_j \in E$ in a VDG $G=(E,\sigma,\rho)$. As explained earlier, positive and negative value-related dependencies can simultaneously exist between a pair of elements $e_i$ and $e_j$. The strength of all positive value-related dependencies from $e_i$ to $e_j$ is denoted by $\rho_{i,j}^{+\infty}$ and calculated by (\ref{Eq_ultimate_strength_positive}), that is to find the strength of the strongest positive value-related dependency~\cite{rosenfeld_fuzzygraph_1975} among all the positive value-related dependencies from $e_i$ to $e_j$. Fuzzy operators $\wedge$ and $\vee$ denote Zadeh's~\cite{zadeh_fuzzysets_1965} fuzzy AND (taking minimum) and fuzzy OR (taking maximum) operations respectively. In a similar way, the strength of all negative value-rrelated dependencies from $e_i$ to $e_j$ is denoted by $\rho_{i,j}^{-\infty}$ and calculated by (\ref{Eq_ultimate_strength_negative}).

\vspace{0.1em}
%-----------------------------------------------------
\begin{align}
\label{Eq_ultimate_strength_positive}
&\rho^{+\infty}_{i,j} = \bigvee_{\substack{d_m \in D \\ \quad\sigma(d_m)=+}} \quad \rho(d_m)  \\[5pt]
\label{Eq_ultimate_strength_negative}
&\rho^{-\infty}_{i,j} = \bigvee_{\substack{d_m \in D \\ \quad\sigma(d_m)=-}} \quad \rho(d_m) \\[5pt]
\label{Eq_influence}
&I_{i,j} = \rho_{i,j}^{+\infty}-\rho_{i,j}^{-\infty} 
\end{align}
%-----------------------------------------------------
\vspace{0.1em}

The overall strength of all positive and negative value-related dependencies from $e_i$ to $e_j$, is referred to as the \textit{Overall Influence} of $e_j$ on the value of $e_i$ and denoted by $I_{i,j}$. As given by (\ref{Eq_influence}), $I_{i,j}\in[-1,1]$ is calculated by subtracting the strength of all negative value-related dependencies ($\rho_{i,j}^{-\infty}$) from the strength of all positive value-related dependencies ($\rho_{i,j}^{+\infty}$). $I_{i,j}>0$ states that $e_j$ will ultimately influence the value of $e_i$ in a positive way whereas $I_{i,j}<0$ indicates that the ultimate influence of $e_j$ on $e_i$ is negative.

\section{The Integer Programming Model}
\label{sew_factoring_dars}

The extent to which the value of an element $e_i$ is influenced by ignoring positive value-related dependencies and/or selecting negative value-related dependencies of $e_i$, is referred to as the penalty~\cite{mougouei2017dasrp} of $e_i$ and denoted by $p_i$. $p_i$ is calculated by taking supremum over the overall influences of all ignored positive dependencies and selected negative dependencies of $e_i$ as given by (\ref{Eq_penalty}). In this equation, $n$ denotes the total number of elements ($E:\{e_1,...,e_n\}$) and $x_j$ specifies whether an elements $e_j$ is selected ($x_j=1$) or not ($x_j=0$). 

\begin{align}
\label{Eq_penalty}
& \text{ }p_{i}= \displaystyle \bigvee_{\substack{j=1 \\ j \neq i}}^{n} \bigg(\frac{\lvert I_{i,j} \rvert + (1-2x_j)I_{i,j}}{2}\bigg),\phantom{s}x_j \in\{0,1\}
\end{align}

%
%\begin{align}
%\label{Eq_penalty}
%& \text{ }p_{i}= \displaystyle \bigvee_{\substack{j=1 \\ j \neq i}}^{n} \bigg(\frac{\lvert I_{i,j} \rvert + (1-2x_j)I_{i,j}}{2}\bigg),\phantom{s}x_j \in\{0,1\}
%\end{align}
%
%For an element $e_i$, (\ref{Eq_value}) derives the expected value of $e_i$ denoted by $v^\prime_i$ by subtracting $p_i v_i$ from the estimated value of $e_i$ ($v_i$). As such, overall value (OV) of an element set $E=\{e_1,...,e_n\}$ can be calculated by accumulating the expected values of selected elements as given by~(\ref{Eq_ocv}). 

Equations (\ref{Eq_dars})-(\ref{Eq_dars_c3}) give our proposed integer linear programming model where $x_i$ is a selection variable denoting whether an element $e_i \in E:\{e_1,...,e_n\}$ is selected ($x_i=1$) or not ($x_i=0$). Also, $w_i$ and $v_i$ denote weight and value of an element $e_i$ respectively while $W$ specifies the capacity of the knapsack. Moreover, $p_i$ denotes the penalty of an element $e_i$ which is the extent to which ignoring positive value-related dependencies and/or selecting negative value-related dependencies of $e_i$ influence the value of $e_i$.  %$p_i$ as given by (\ref{Eq_penalty}) is calculated through taking supremum over the influences of all ignored positive value-related dependencies and selected negative value-related dependencies of $e_i$. This is given by (\ref{Eq_dars_c2}).

\begin{align}
\label{Eq_dars}
\text{Maximize } & \sum_{i=1}^{n} x_i (1-p_i) v_i\\[5pt]
\label{Eq_dars_c1}
\text{Subject to} & \sum_{i=1}^{n} w_i x_i \leq W\\[5pt] 
\label{Eq_dars_c2}
& p_i \ge \displaystyle \bigg(\frac{\lvert I_{i,j} \rvert + (1-2x_j)I_{i,j}}{2}\bigg),&\quad i\neq j = 1,...,n\\[5pt]
\label{Eq_dars_c3}
&x_i \in \{0,1\},&\quad i = 1,...,n
\end{align}

%\begin{align}
%%\label{Eq_cs}
%%& \phi_i = 1-p_i\\
%\label{Eq_value}
%& v^\prime_i = (1-p_i)v_i \\
%\label{Eq_ocv}
%&OV = \sum_{i=1}^{n} x_i (1-p_i)v_i, \textit{ } x_i \in \{0,1\}
%\end{align}

For an element $e_i$, $p_i$ is a function of $x_j$ and $I_{i,j}$ that is $p_i=f(x_j,I_{i,j})$. Hence, objective function of (\ref{Eq_dars}) can be reformulated as $\sum_{i=1}^{n} x_i v_i - f(x_j,I_{i,j})x_iv_i$ with a quadratic non-linear expression~\cite{burkard_1984_quadratic} $exp=x_i f(x_j,I_{i,j})v_i$ which shows the integer programming model (\ref{Eq_dars})-(\ref{Eq_dars_c3}) is a non-linear model which cannot be solved efficiently nor can it be approximated efficiently~\cite{burkard_1984_quadratic,sahni_1976_quadretic}. Integer linear programming models on the other hand, are known to be solvable efficiently due to the advances in tools~\cite{veerapen2015integer}. 
%Moreover, Veerapen\textit{et al.} recently demonstrated that large scale single objective release planning problems (such as the DA-SRP) can be very efficiently solved using integer linear programming~\cite{veerapen2015integer}. 

Hence, we have converted (\ref{Eq_dars}) to its corresponding linear form by substituting the non-linear expression $x_ip_i$ with the linear expression $y_i$ ($y_i=x_ip_i$). As such, either $a:(x_i=0) \rightarrow (y_i=0)$, or $b:(x_i=1) \rightarrow (y_i=p_i)$ can occur. In order to capture the relation between $p_i$ and $y_i$ in a linear form, we have introduced an auxiliary variable $g_i=\{0,1\}$ and Constraints (\ref{Eq_dars_linear_c3})-(\ref{Eq_dars_linear_c7}) are added to the initial model. As such, we have either $(g_i=0) \rightarrow a$, or $(g_i=1) \rightarrow b$. The resulting model as given by (\ref{Eq_dars_linear})-(\ref{Eq_dars_linear_c7}) is an integer linear model.

\begin{align}
\label{Eq_dars_linear}
&\text{Maximize }  \sum_{i=1}^{n} x_i v_i - y_i v_i\\[5pt]
\label{Eq_dars_linear_c1}
&\text{Subject to} \sum_{i=1}^{n} w_i x_i \leq W\\[5pt]
\label{Eq_dars_linear_c2}
& p_i \ge \displaystyle \bigg(\frac{\lvert I_{i,j} \rvert + (1-2x_j)I_{i,j}}{2}\bigg),\quad& i\neq j = 1,...,n\\[5pt]
\label{Eq_dars_linear_c3}
& -g_i \leq x_i \leq  g_i,&\quad i=1,...,n\\[5pt]
\label{Eq_dars_linear_c4}
& 1-(1-g_i) \leq x_i \leq 1+(1-g_i),&\quad i=1,...,n\\[5pt]
\label{Eq_dars_linear_c5}
& -g_i \leq y_i \leq g_i,&\quad i=1,...,n\\[5pt]
\label{Eq_dars_linear_c6}
& -(1-g_i)\leq(y_i-p_i) \leq (1-g_i),&\quad i=1,...,n\\[5pt]
\label{Eq_dars_linear_c7}
& \text{ }x_i,y_i,g_i \in \{0,1\},&\quad i=1,...,n
\end{align}

\vspace{1em}
% Example was commented out as we r presenting a case study and btw example can not show how selection 
%\input{validation}
%\input{validation_case}
%\vspace{2em}
%\input{validation_simulation}
%\input{validation_simulation_design}
%\input{validation_simulation_result}
%\input{scalability}
%\input{limitation}
\section{Summary}
\label{sec_conclusion}

In this paper we presented an integer programming model for solving a variation of binary knapsack problem where value of each element may be influenced by selecting or ignoring other elements. The strengths of these influences (value-related dependencies) are assumed to be imprecise and hard to express. Algebraic structure of fuzzy graphs was used for capturing the imprecision associated with strengths of value-related dependencies. 

\bibliographystyle{abbrv}
\bibliography{ref} 

\end{document}